\documentclass[letterpaper, 10 pt, conference]{ieeeconf}

\IEEEoverridecommandlockouts                              

\usepackage{cite}
\usepackage{amsmath,amssymb,amsfonts}
\usepackage{algorithmic}
\usepackage{graphicx}
\usepackage{textcomp}
\usepackage{xcolor}
\usepackage{tikz}
\usetikzlibrary{arrows.meta,positioning,fit,backgrounds,shapes.geometric, calc, shapes.symbols}
\usepackage{acronym}

\usepackage[hidelinks]{hyperref}

\acrodef{ROS}{Robot Operating System}
\acrodef{DAG}{Directed Acyclic Graph}
\acrodef{CI}{Continuous Integration}
\acrodef{ML}{Machine Learning}
\acrodef{xattr}{extended attribute}

\newcommand{\Fig}[1]{Fig.~\ref{#1}}
\newcommand{\Sec}[1]{Sec.~\ref{#1}}

\newcommand{\Tab}[1]{Table~\ref{#1}}

\title{\LARGE \bf
Modeling Robotics Dataset Construction as an Artifact-Based Build Process
}

\author{
Leon Pohl, Lukas Beer, George Sebastian, and Mirko Maehlisch%
\thanks{All authors are with the Institute for Autonomous Driving, University of the Bundeswehr Munich, 85579 Neubiberg, Germany.\protect\\
Corresponding author: Leon Pohl, \texttt{leon.pohl@unibw.de}.\protect\\
Code: \protect\url{https://github.com/UniBwTAS/bagzel}.}
}

\let\arxivOriginalThanks\thanks
\renewcommand{\thanks}[1]{%
  \arxivOriginalThanks{#1\protect\\
  DOI: \protect\url{https://doi.org/10.1109/CASE69030.2026.11704392}\protect\rlap{.}}%
}

\newcommand{\ArxivTexheightBalance}{-0.0cm}

\usepackage{eso-pic}

\newcommand{\AddArxivHeader}{%
\AddToShipoutPictureFG*{%
  \AtPageUpperLeft{%
    \hspace{0.65in}%
    \raisebox{-0.43in}{%
      \parbox{7.2in}{%
        \scriptsize 2026 IEEE 22nd International Conference on Automation Science and Engineering (CASE 2026). Accepted Author Manuscript, October 2026.
      }%
    }%
  }%
}%
}

\newcommand{\AddIEEECopyrightFooter}{%
\AddToShipoutPictureFG*{%
  \AtPageLowerLeft{%
    \hspace{0.65in}%
    \raisebox{0.35in}{%
      \parbox{7.2in}{%
        \scriptsize
        \textcopyright{} 2026 IEEE. Personal use of this material is permitted.
        Permission from IEEE must be obtained for all other uses, in any current
        or future media, including reprinting/republishing this material for
        advertising or promotional purposes, creating new collective works, for
        resale or redistribution to servers or lists, or reuse of any copyrighted
        component of this work in other works.
      }%
    }%
  }%
}%
}

\begin{document}

\AddArxivHeader
\AddIEEECopyrightFooter


\bstctlcite{bibcontrol_IEEE_etal6}

\maketitle
\thispagestyle{empty}
\pagestyle{empty}

\begin{abstract}

Robotic systems generate large volumes of multimodal sensor data, but converting ROS bag recordings into machine learning datasets is often handled by ad hoc sequential scripts, creating engineering overhead and slow iteration cycles. We model dataset construction as an artifact-based build process over a dependency graph and implement this approach in Bagzel, an open-source Bazel extension for reproducible, incremental dataset generation (including nuScenes-format export). We compare Bagzel and Bagzel-xattr (server-side digest management) against a sequential rosbag2nuscenes baseline. Bagzel reduces runtime in all evaluated execution modes, with the largest gains in iterative workflows (up to \textbf{386.26$\times$} in warm builds and \textbf{7.21$\times$} in incremental builds on a 20.4\,GB dataset). Across dataset sizes from 5.1 to 20.4\,GB, Bagzel variants show markedly better scaling behavior than the baseline, especially in warm and incremental modes. Bagzel-xattr provides additional gains, with a mean runtime reduction of 5.9\% compared to Bagzel in the input granularity study. Overall, modeling robotics dataset construction as an artifact-based build process substantially reduces dataset update latency while maintaining a deterministic build design that supports reproducibility.

\end{abstract}

\acresetall
\section{Introduction}

Modern robotics research increasingly relies on supervised and self-supervised learning methods that require large and diverse datasets for training and evaluation \cite{brohan2023rt1,zitkovich2023rt2,oneill2024ope}. In autonomous driving in particular, public benchmarks such as nuScenes \cite{caesar2020nus}, KITTI \cite{geiger2012are}, Waymo Open Dataset \cite{sun2020sca}, and Argoverse 2 \cite{wilson2021arg} have accelerated progress and become de facto standards, shaping model architectures and enabling reproducible comparison.

However, the widespread adoption of these benchmarks has led to an implicit coupling between research codebases and dataset-specific formats. Many learning architectures and evaluation pipelines are implemented directly against benchmark schemas \cite{gao2024vis,zhangResWorld2026a,li2025end}, meaning that newly recorded data must either be converted into these formats or require adaptation of the training code. This coupling increases engineering overhead and complicates experimentation with custom datasets. It also constrains evaluation beyond fixed benchmark distributions, where performance can degrade under even small distribution shifts \cite{recht2019ima}. In active learning settings, where models are iteratively retrained on newly collected edge-case data, such integration friction directly slows down the training loop. This motivates the development of streamlined dataset construction workflows capable of matching the fast cycles of model development and deployment.

In robotics, new data are commonly captured in the \ac{ROS} ecosystem \cite{quigley2009ros,macenski2022rob} as rosbag recordings that store synchronized multi-sensor streams and system state. Although well suited for replay and debugging, rosbags are not directly organized as machine learning datasets and require substantial post-processing before training or evaluation. As observed in prior work on \ac{ML} pipelines \cite{sculley2015hid}, such ad hoc integration can accumulate technical debt, reduce reproducibility, and increase operational cost of \ac{ML} methods at scale. More broadly, production \ac{ML} systems face persistent data-management challenges in data validation, evolution, and maintenance \cite{polyzotis2017dat}. 

\begin{figure}
\centering
\resizebox{0.98\columnwidth}{!}{%
\begin{tikzpicture}[
  >=Latex,
  line cap=round,
  line join=round,
  flow/.style={->, line width=0.95pt, draw=black!85},
  dflow/.style={->, dashed, line width=0.95pt, draw=black!70},
  lbl/.style={font=\scriptsize, text=black!80},
  build/.style={draw=orange!55!black, rounded corners=2pt, fill=orange!12, align=center, font=\scriptsize, minimum width=22mm, minimum height=10mm},
  cache/.style={draw=gray!60, rounded corners=2pt, fill=gray!12, align=center, font=\scriptsize, minimum width=16mm, minimum height=7mm},
  data/.style={draw=green!35!black, cylinder, shape border rotate=90, aspect=0.25,
               fill=green!12, minimum height=7mm, minimum width=18mm, align=center, font=\scriptsize}
]

\begin{scope}[shift={(-0.65,-0.35)}]
  \draw[fill=blue!30, draw=black!70, line width=0.6pt] (-0.35,0.05) rectangle (0.35,0.22);
  \draw[fill=blue!30, draw=black!70, line width=0.6pt] (-0.12,0.22) -- (0.01,0.38) -- (0.24,0.38) -- (0.35,0.22) -- cycle;
  \draw[fill=black!80] (-0.18,0.01) circle (0.045);
  \draw[fill=black!80] (0.17,0.01) circle (0.045);
  \node[lbl] at (0,-0.275) {Cars};
\end{scope}

\begin{scope}[shift={(0.65,-0.30)}]
  \draw[fill=gray!20, draw=black!70, rounded corners=1pt, line width=0.6pt] (-0.30,-0.20) rectangle (0.30,0.20);
  \draw[fill=gray!15, draw=black!70, line width=0.6pt] (-0.06,0.20) rectangle (0.06,0.32);
  \draw[draw=black!70, line width=0.6pt] (0.00,0.32) -- (0.00,0.45);
  \draw[fill=black!80] (0.00,0.48) circle (0.018);
  \draw[fill=black!80] (-0.16,0.00) circle (0.025);
  \draw[fill=black!80] (0.14,0.00) circle (0.025);
  \node[lbl] at (0,-0.35) {Robots};
\end{scope}

\node[cloud, draw=cyan!40!black, fill=cyan!15,
      cloud puffs=9, cloud puff arc=100,
      aspect=1.6, minimum width=14mm, minimum height=8mm,
      font=\scriptsize, inner sep=1pt] (cloud) at (0,0.95) {File Server};

\node[build] (build) at (2.35,2.0) {Build System\\(Deterministic \\\& Reproducible)};
\node[cache] (cache) at (2.35,0.65) {Artifact Cache};
\node[data] (dataset) at (4.75,0.60) {Machine\\Learning Dataset};

\draw[flow] (-0.55,0.1) -- node[lbl,left] {raw data} (-0.18,0.56);
\draw[flow] (0.55,0.1) -- node[lbl,right] {raw data} (0.10,0.56);

\draw[flow] (cloud.north) |- node[lbl,pos=0.45,above=2pt] {hashes} (build.west); 

\draw[dflow] ([xshift=-1.5mm]build.south) -- node[lbl,left]  {stores} ([xshift=-1.5mm]cache.north);
\draw[dflow] ([xshift= 1.5mm]cache.north) -- node[lbl,right] {reuses}           ([xshift= 1.5mm]build.south);

\draw[flow] (build.east) -| node[lbl,pos=0.45,above=0pt] {incremental updates} (dataset.north);

\end{tikzpicture}%
}
\caption{Artifact-based robotics dataset construction workflow. Raw data from cars and robots are uploaded to a file server, whose content hashes trigger a deterministic build system. The build system stores and reuses intermediate artifacts via the cache and incrementally generates \ac{ML} datasets.}
\label{fig:bagzel-build-cache}
\end{figure}
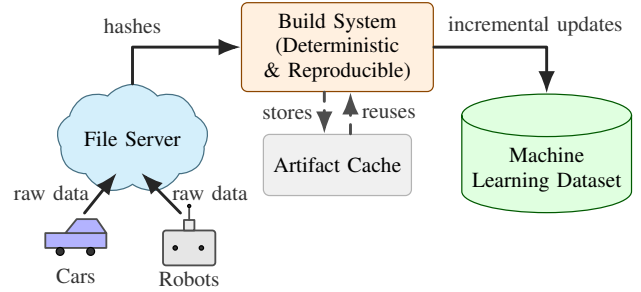

Similar scalability challenges have been reported in large \ac{CI} pipelines, where long dependency chains can increase feedback latency and reduce developer productivity \cite{bernardo2023imp}. A common mitigation strategy is to use artifact-based build systems, such as Bazel \cite{bazelteam2025baz}, Buck \cite{buck2013}, and Pants \cite{pantsbuild}. These systems model the build as a \ac{DAG}, in which nodes represent build steps and edges represent declared dependencies. For each step, the required inputs and produced outputs are specified explicitly and tracked via hashing. This allows the build system to re-execute only affected steps and reuse cached artifacts when inputs are unchanged. With these principles implemented in Bazel CI pipelines, reported build-time speedups reach up to 12.8$\times$ \cite{zheng2024doe}.

We transfer this execution principle to robotics dataset construction. The contributions of this work are as follows:

\begin{itemize}
\item We formulate robotics dataset construction as an artifact-based build problem by modeling preprocessing stages as explicit build targets with declared inputs and outputs. This enables dependency tracking, selective recomputation, and artifact reuse for dataset generation, as illustrated in \Fig{fig:bagzel-build-cache}.

\item We design and implement \textbf{Bagzel} \cite{pohl2026bag}, an open-source framework built on the Bazel build system that realizes this formulation and enables reproducible and incremental generation of nuScenes-format datasets from \ac{ROS} recordings.

\item We introduce a server-side digest management mechanism for large input files (\textbf{Bagzel-xattr}) that reduces hashing overhead during the build process while preserving deterministic artifact identification.
\end{itemize}

We empirically evaluate the proposed approach with respect to execution efficiency, scalability with increasing dataset size, and the influence of input granularity and server-side digest management on incremental rebuild performance.

\section{Related Work}

Recent ROS ecosystem reports indicate strong \ac{ROS}~2 adoption and continued community growth, increasing demand for scalable and reproducible preprocessing \cite{ros_metrics_2025}. Building on ROS logging infrastructure \cite{quigley2009ros,macenski2022rob}, prior work addresses specific stages of dataset creation, including rosbag conversion \cite{chrosniak2023ros, kulkarni2023rac}, synchronization validation \cite{vigni2024ros}, and LLM-assisted annotation \cite{zhang2025ros}.

Beyond robotics converters, MLOps literature emphasizes dataset and model versioning because \ac{ML} performance depends strongly on data quality. Tools such as DVC \cite{dvc}, MLflow \cite{chen2020deva}, and TFX \cite{baylor2017tfxa} support experiment tracking and artifact management, and recent studies analyze practical versioning strategies for \ac{ML} data and models \cite{barretosimedopacheco2024dvc, barrak2021coe}. In contrast, this work formulates robotics dataset construction as an artifact-based build process, applying core \ac{CI} principles (explicit dependencies, hermetic actions, content-addressed caching, incremental rebuilds) to support deterministic, low-latency updates and integration into existing \ac{CI} workflows.

\section{Methodology}

We formulate dataset construction from raw recordings as an artifact-based build problem over a \ac{DAG}. This formulation underlies Bagzel. We use an abstract bipartite build graph $\mathcal{P}=(V,E)$ with node set $V = V_a \cup V_o$ and edge set $E$, where $V_a$ denotes artifact nodes (data objects) and $V_o$ denotes operation nodes (transformations), with $V_a \cap V_o = \emptyset$. Edges encode declared dependencies and alternate between artifacts and operations, i.e.,
\begin{equation}
E \subseteq (V_a \times V_o) \cup (V_o \times V_a).
\end{equation}
Thus, artifacts serve as inputs to operations, and operations produce derived artifacts. We denote source artifacts by $s_i \in V_a$, operations by $o_i \in V_o$, and derived artifacts by $a_i \in V_a$. 

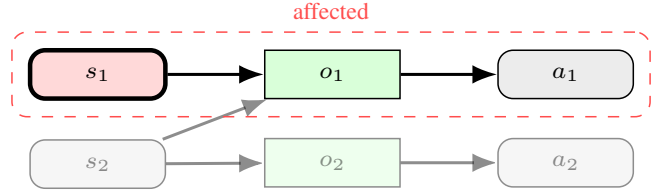
\begin{figure}
\centering
\resizebox{\linewidth}{!}{%
\begin{tikzpicture}[
  scale=0.9, transform shape,
  node distance=4mm and 6mm,
  >=Latex,
  artifact/.style={draw, rounded corners, fill=gray!15, minimum width=14mm, minimum height=5mm, align=center, font=\scriptsize},
  op/.style={draw, rectangle, fill=green!15, minimum width=14mm, minimum height=5mm, align=center, font=\scriptsize},
  changed/.style={draw, rounded corners, fill=red!15, very thick, minimum width=14mm, minimum height=5mm, align=center, font=\scriptsize},
  faded/.style={opacity=0.45},
  edge/.style={->, thick}
]
\node[changed] (s1) {$s_1$};
\node[artifact, below=of s1, faded] (s2) {$s_2$};

\node[op, right=10mm of s1] (o1) {$o_1$};
\node[op, below=of o1, faded] (o2) {$o_2$};

\node[artifact, right=10mm of o1] (a1) {$a_1$};
\node[artifact, below=of a1, faded] (a2) {$a_2$};

\draw[edge] (s1) -- (o1);
\draw[edge, faded] (s2) -- (o1);
\draw[edge, faded] (s2) -- (o2);
\draw[edge] (o1) -- (a1);
\draw[edge, faded] (o2) -- (a2);

\node[
  draw=red!70, dashed, rounded corners, inner sep=1.5mm,
  fit=(s1)(o1)(a1),
  label={[red!70, font=\scriptsize]above:affected}
] {};
\end{tikzpicture}%
}
\caption{Abstract artifact \ac{DAG} $\mathcal{P}=(V,E)$ for selective recomputation. A modification to $s_1$ propagates along dependency edges and invalidates only the affected path ($s_1 \to o_1 \to a_1$), while unrelated nodes ($s_2$, $o_2$, and $a_2$) remain reusable.}
\label{fig:artifact-dag}
\end{figure}

Each operation declares its inputs, outputs, and execution rule. Rebuild decisions are based on an action digest computed from the declared inputs and operation definition. If the digest changes, the operation and its transitive dependents are recomputed (\Fig{fig:artifact-dag}); otherwise, cached artifacts are reused. Under deterministic and hermetic execution, identical inputs and rules yield identical digests and outputs, enabling correct cache reuse. The graph-processing cost of an incremental update is linear in the size of the affected subgraph, $O(|V_\Delta| + |E_\Delta|)$. Let $V_{o,\Delta} \subseteq V_o$ denote the recomputed operation nodes. The total runtime additionally includes their execution cost, $\sum_{o \in V_{o,\Delta}} T(o)$.

Artifact-based build systems are typically optimized for many small source files, where client-side content hashing is inexpensive. In robotics data pipelines, raw recordings can reach hundreds of gigabytes, making repeated full-file hashing on the client a bottleneck, particularly during iterative updates with only small changes.

To reduce this overhead, we extend the base formulation with server-side digest management, implemented in Bagzel-xattr. When a source artifact changes, a detection service recomputes and persists its digest as metadata associated with that artifact. During build graph evaluation, the system uses the stored digest to determine whether dependent steps must be executed, avoiding repeated client-side full-file hashing.

\begin{figure}
\centering
\resizebox{\linewidth}{!}{%
\begin{tikzpicture}[
  scale=0.9, transform shape,
  node distance=5mm and 10mm,
  >=Latex,
  src/.style={draw, rounded corners, fill=gray!15, minimum width=14mm, minimum height=5mm, align=center, font=\scriptsize},
  op/.style={draw, rectangle, fill=green!15, minimum width=14mm, minimum height=5mm, align=center, font=\scriptsize},
  art/.style={draw, rounded corners, fill=gray!15, minimum width=15mm, minimum height=5mm, align=center, font=\scriptsize},
  dec/.style={draw, diamond, aspect=1.5, fill=orange!20, align=center, inner xsep=0.5mm, inner ysep=0.5mm, font=\scriptsize},
  edge/.style={->, thick, black},                 
  ctrl/.style={->, thick, blue!70},               
  meta/.style={->, dashed, thick, blue!70},       
  skip/.style={->, thin, gray!70}                 
]

\node[src] (s1) {$s_1$};
\node[op, right=10mm of s1] (o1) {$o_1$};
\node[art, right=10mm of o1] (a1) {$a_1$};

\draw[edge] (s1) -- (o1);
\draw[edge] (o1) -- (a1);

\node[dec, above=10mm of s1] (chg) {source\\modified?};
\node[op, right=10mm of chg] (oh) {$o_{\mathrm{hash}}$};
\node[art, right=10mm of oh] (ax) {$a_{\mathrm{digest}}$};

\draw[ctrl] (s1.north) -- node[left, font=\scriptsize]{check} (chg.south);
\draw[ctrl] (chg.east) -- node[above, font=\scriptsize]{yes} (oh.west);
\draw[ctrl] (oh) -- (ax);

\draw[skip]
  (chg.north west) to[out=140,in=220,looseness=6]
  node[left, font=\scriptsize, text=gray!70]{no}
  (chg.west);

\draw[meta]
  (ax.south) -- ++(0,-9mm)
  -- node[midway, above=-0.075, font=\scriptsize]{attach digest} ++(-44mm,0)
  -- (s1.north east);

\node[
  draw=gray!70, rounded corners, dashed, inner sep=2mm,
  fit=(chg)(oh)(ax),
  label={[font=\scriptsize]above:File Server}
] {};

\end{tikzpicture}%
}
\caption{Server-side digest management used by Bagzel-xattr. The file server checks whether the source artifact has changed; on a positive decision, it computes and stores a digest attached to $s_1$. Operation $o_1$ then uses this digest for build decisions.}
\label{fig:large-file-hashing}
\end{figure}
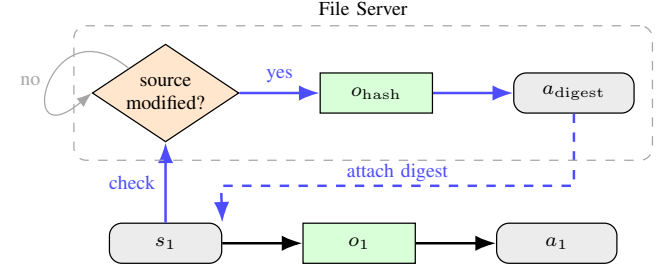

This approach preserves deterministic artifact identification while reducing incremental update overhead for large inputs. \Fig{fig:large-file-hashing} illustrates the change detection flow, digest update path, and subsequent use of stored digests in the build.

\section{Implementation}

To validate the concept of artifact-based dataset construction, we implemented our approach on top of the open-source build system Bazel \cite{bazelteam2025baz}. Bazel was selected due to its explicit modeling of artifacts, deterministic execution model, and content-addressable caching, which provide strong guarantees for reproducibility and incremental recomputation. 

The current implementation supports \ac{ROS}~1 and \ac{ROS}~2 bag files as primary source artifacts. To integrate \ac{ROS} bag processing into the Bazel build model, we developed custom rules that encapsulate common dataset preparation steps, including frame decoding, trajectory extraction, annotation processing, and export to standardized dataset formats. Each rule explicitly declares its inputs, such as raw bag files, and outputs, such as processed dataset artifacts, thereby incorporating dataset construction into the Bazel dependency graph. These extensions are provided as an open-source Bazel rule set named Bagzel.

Bagzel currently supports the generation of two dataset formats: (i) a visual dataset format that organizes sensor data and annotations in a structured directory layout accompanied by metadata files, and (ii) the widely adopted nuScenes format \cite{caesar2020nus}, which facilitates compatibility with established tools and benchmarks in autonomous driving research. Bagzel also supports integration with the Slurm workload manager \cite{yoo2003slu} for distributed execution of independent build actions across cluster nodes.

We evaluate two Bagzel configurations and one sequential baseline. The default Bagzel configuration performs client-side build-time hashing, where digests of input artifacts are computed during each build. The Bagzel-xattr variant extends Bagzel with server-side digest management, in which file digests are maintained on the file server and reused during build evaluation. As a baseline, we adapt the sequential converter rosbag2nuscenes \cite{chrosniak2023ros,kulkarni2023rac}, which follows a monolithic, single-file processing pipeline and converts one \ac{ROS} bag input at a time into nuScenes outputs. The baseline does not implement artifact caching or selective invalidation across processing steps.

\section{Experimental Setup}

To empirically evaluate our contributions and understand their practical impact on robotics dataset construction, we investigate the following research questions:
\begin{itemize}
\item \textbf{RQ1 (Execution Efficiency):}
Does artifact-based dataset construction reduce overall and iterative runtime compared to a conventional sequential pipeline?
\item \textbf{RQ2 (Scalability):}
How does runtime scale with increasing dataset size under artifact-based versus sequential dataset construction?
\item \textbf{RQ3 (Input Granularity and Digest Management):}
How do input granularity and server-side digest management influence incremental runtime?
\end{itemize}

To answer these questions, we design two groups of experiments. RQ1 and RQ2 are addressed by comparing Bagzel, Bagzel-xattr, and rosbag2nuscenes across execution modes and dataset sizes. RQ3 is addressed by a separate input granularity ablation that compares the incremental update behavior of Bagzel and Bagzel-xattr under fixed total data volume.

The scalability experiments were conducted using three datasets of increasing size: small (5.1\,GB), medium (10.2\,GB), and large (20.4\,GB). The small dataset consists of one \ac{ROS}~1 bag file (2.5\,GB) and one \ac{ROS}~2 bag file (2.6\,GB) containing identical sensor data. The \ac{ROS}~1 bag was converted to \ac{ROS}~2 format to ensure functional equivalence between formats while maintaining separate input artifacts. The medium dataset was constructed by duplicating the small dataset, resulting in two \ac{ROS}~1 and two \ac{ROS}~2 bag files with a total size of 10.2\,GB. The large dataset extends this setup by further duplication, yielding four \ac{ROS}~1 and four \ac{ROS}~2 bag files with a total size of 20.4\,GB. The execution efficiency experiments were performed on the large dataset.

The primary performance metric is end-to-end wall-clock runtime \(T\) per build execution. Based on these measurements, we compute (i) relative speedup and percentage runtime reduction with respect to the baseline for each execution mode, and (ii) runtime scaling behavior as a function of dataset size using a first-order linear fit over the measured range.

Runtime reduction and speedup are defined as
\begin{align}
\mathrm{Reduction}(\%) &=
\left(1-\frac{T_{\mathrm{method}}}{T_{\mathrm{base}}}\right)\times 100,\\
\mathrm{Speedup} &=
\frac{T_{\mathrm{base}}}{T_{\mathrm{method}}}.
\end{align}
where \(T_{\mathrm{base}}\) denotes the baseline and \(T_{\mathrm{method}}\) the evaluated method runtime.

To characterize scalability, we model runtime as a linear function of dataset size \(D\) (in GB)
\begin{equation}
T(D) = aD + b,
\label{eq:linear-scaling}
\end{equation}
where \(T(D)\) denotes runtime in seconds, \(a\) (s/GB) represents the slope of runtime growth with respect to dataset size, and \(b\) is the intercept of the linear model.

We evaluate three execution modes under identical build conditions: (i) \textbf{Cold build}, where no reusable cache entries or prepared build environment are available and all actions are executed; (ii) \textbf{Warm build}, where no input changes are introduced and previously generated artifacts and build environment are reused from cache; and (iii) \textbf{Incremental build}, where a localized input modification is introduced and only the affected subgraph is rebuilt. Each configuration was executed in three independent trials, and the median runtime is reported.

For incremental rebuild experiments, we introduced a metadata change in one source artifact. This modifies the file digest without changing the sensor content used by the extraction step, producing a controlled and localized invalidation in the dependency graph. As a result, the incremental recomputation workload should remain comparable across dataset sizes while simulating a small data update.

In the input granularity ablation, we keep the base dataset size constant at 20.4\,GB and vary only how it is partitioned (one, two, or eight bag files). A separate rebuild target \(\mathrm{R}\) (2.5\,GB) is included in every condition and is identical across runs. Thus, each condition consists of a fixed 20.4\,GB base dataset plus the same rebuild target \(\mathrm{R}\) (2.5\,GB). The conditions are: (i) \textbf{one input file} + rebuild target \(\mathrm{R}\), (ii) \textbf{two input files} + rebuild target \(\mathrm{R}\), and (iii) \textbf{eight input files} + rebuild target \(\mathrm{R}\). We only compare Bagzel and Bagzel-xattr in this experiment and compare the incremental runtime of the modified target \(\mathrm{R}\) across conditions. This setup isolates the effect of input granularity on hashing and incremental runtime while controlling for total base data volume. 

All experiments were conducted on a workstation equipped with an 11th Gen Intel Core i7-11700K CPU (8 cores, 16 threads, 3.6\,GHz base frequency), 64\,GB RAM, and a 4\,TB Samsung SSD 990 PRO NVMe. The system ran Ubuntu 20.04.6 LTS with Linux kernel 5.15. Bazel version 8.3.1 was used for all builds. No GPU acceleration was employed. To ensure a fair comparison, all experiments were executed under identical system conditions. Remote caching was disabled, and all experiments were performed locally on the workstation.

\section{Results}

This section reports results by research question (RQ). \Sec{sec:rq1} compares execution efficiency against the sequential baseline, \Sec{sec:rq2} analyzes scaling with dataset size, and \Sec{sec:rq3} evaluates the effect of input artifact granularity on incremental build performance.

\subsection{RQ1: Execution Efficiency} \label{sec:rq1}
\begin{figure}
\centering
\includegraphics[scale=1.0]{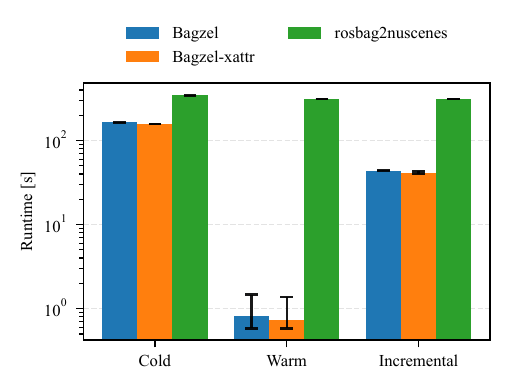}
\caption{Runtime comparison across execution modes for the large dataset (20.4\,GB). Bars show the median of three runs; error bars indicate the min-max range. The y-axis is logarithmic.}
\label{fig:build-bar-large}
\end{figure}

\Fig{fig:build-bar-large} reports runtimes for the large dataset by execution mode. Bars show median runtime over three trials. Error bars indicate min-max range. Bagzel and Bagzel-xattr outperform the sequential baseline in all modes, with the largest gains in warm and incremental builds. The Bagzel-xattr variant provides an additional incremental benefit over default Bagzel. Because the y-axis is logarithmic, uncertainty should be interpreted in relative terms. Consequently, warm mode error bars may appear visually large despite modest absolute runtime differences.

The numerical runtime results are summarized in \Tab{tab:runtime-large}. Compared with rosbag2nuscenes, Bagzel substantially improves performance across all build modes. For cold builds, runtime decreases from 343.40\,s to 166.40\,s (51.5\% reduction, \(2.06\times\) speedup). The improvement is most pronounced in warm builds, where runtime drops from 312.87\,s to 0.81\,s, corresponding to a 99.7\% reduction (\(386.26\times\)). For incremental builds, runtime decreases from 313.11\,s to 43.45\,s (86.1\% reduction, \(7.21\times\)).

Bagzel-xattr further improves performance over Bagzel with modest but consistent gains. Runtime decreases from 166.40\,s to 156.96\,s in cold builds (5.7\%, \(1.06\times\)), from 0.81\,s to 0.74\,s in warm builds (8.6\%, \(1.10\times\)), and from 43.45\,s to 41.34\,s in incremental builds (4.9\%, \(1.05\times\)).

\begin{table}
\vspace{2mm}
\centering
\caption{Large dataset runtime comparison (median of three runs). R2N = rosbag2nuscenes, B = Bagzel, BX = Bagzel-xattr.}
\label{tab:runtime-large}
\setlength{\tabcolsep}{3.2pt}
\renewcommand{\arraystretch}{0.95}
\begin{tabular}{lccc}
\hline
\textbf{Metric} & \textbf{Cold} & \textbf{Warm} & \textbf{Incremental} \\
\hline
\multicolumn{4}{l}{\textit{Runtime [s]}} \\
R2N & 343.40 & 312.87 & 313.11 \\
B   & 166.40 & 0.81   & 43.45 \\
BX  & 156.96 & 0.74   & 41.34 \\
\hline
\multicolumn{4}{l}{\textit{Speedup [$\times$]}} \\
B vs R2N  & 2.06 & 386.26 & 7.21 \\
BX vs B   & 1.06 & 1.10   & 1.05 \\
\hline
\multicolumn{4}{l}{\textit{Reduction [\%]}} \\
B vs R2N  & 51.5 & 99.7 & 86.1 \\
BX vs B   & 5.7  & 8.6  & 4.9 \\
\hline
\end{tabular}
\end{table}

Overall, the results indicate that artifact-based dataset construction with Bagzel substantially reduces runtime compared to a sequential baseline, particularly in warm and incremental execution modes. The additional optimization of server-side digest management in Bagzel-xattr provides further performance benefits, albeit with more modest gains.

\subsection{RQ2: Scalability} \label{sec:rq2}

\begin{figure*}
\centering
\includegraphics[scale=1.0]{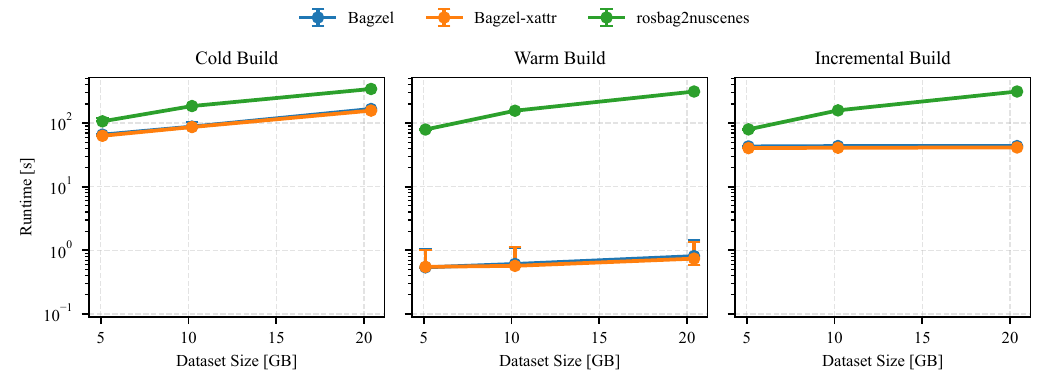}
\caption{Runtime scaling across dataset sizes (5.1, 10.2, and 20.4\,GB) for cold, warm, and incremental execution modes. Points show median runtime over three runs; error bars indicate min-max range. The y-axis is logarithmic.}
\label{fig:build-scaling}
\end{figure*}

To evaluate how performance scales with dataset size, we additionally assessed all methods on a small (5.1\,GB) and a medium (10.2\,GB) dataset. The results are presented in \Fig{fig:build-scaling}, where the x-axis represents dataset size in GB and the y-axis represents runtime in seconds. Separate plots are shown for cold, warm, and incremental build modes. Each method is distinguished by a different color.

The results indicate that all Bagzel variants consistently outperform the rosbag2nuscenes baseline across all three build modes. In the cold build mode, the runtime of all methods increases with dataset size, demonstrating a clear scaling effect. In the warm build mode, the baseline runtime continues to increase with dataset size, whereas Bagzel runtimes remain close to one second.

For incremental rebuilds, the runtime of the Bagzel variants remains nearly constant and shows little dependence on dataset size. In contrast, the baseline runtime continues to increase with dataset size.

We summarize scaling over the measured range (5.1 to 20.4\,GB) using a first-order linear fit  \(T(D) = aD + b\), where \(D\) is the dataset size in GB, \(T(D)\) is the runtime in seconds, and \(a\) is the slope representing the increase in runtime per GB. The fitted slopes (\Tab{tab:scaling-slopes}) show that both Bagzel variants have substantially lower scaling rates than rosbag2nuscenes across all modes (e.g., cold: 6.74/6.25 vs 15.50 s/GB). In warm and incremental modes, Bagzel-based slopes remain close to zero, indicating weak dependence on dataset size within the evaluated range. The small Bagzel vs Bagzel-xattr slope differences (e.g., incremental: 0.03 vs 0.06 s/GB) are minor and should be interpreted cautiously given the limited number of dataset sizes. 

\begin{table}
\centering
\caption{Estimated runtime scaling slopes from linear fits \(T(D)=aD+b\), where \(a\) is reported in s/GB.}
\label{tab:scaling-slopes}
\begin{tabular}{lccc}
\hline
\textbf{Method} & \textbf{Cold} & \textbf{Warm} & \textbf{Incremental} \\
\hline
Bagzel & 6.74 & 0.02 & 0.03 \\
Bagzel-xattr & 6.25 & 0.01 & 0.06 \\
rosbag2nuscenes & 15.50 & 15.29 & 15.25 \\
\hline
\end{tabular}
\end{table}

Overall, within the evaluated range, Bagzel exhibits substantially better scaling behavior than the sequential baseline across all three modes. For both Bagzel variants, runtime in warm and incremental modes shows only weak dependence on dataset size.

\subsection{RQ3: Input Granularity and Digest Management} \label{sec:rq3}

\begin{figure}
\centering
\includegraphics[scale=1.0]{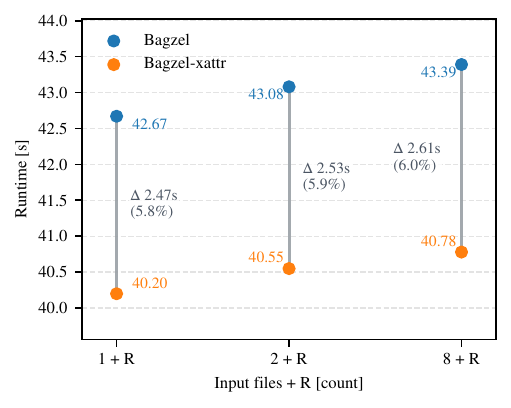}
\caption{Incremental runtime for constant target \(\mathrm{R}\) (2.5\,GB) with a fixed base dataset (20.4\,GB), varying only input file granularity (1+\(\mathrm{R}\), 2+\(\mathrm{R}\), 8+\(\mathrm{R}\)). Points show the median over three runs; connectors indicate the difference between Bagzel and Bagzel-xattr (seconds and \%).}
\label{fig:input-granularity}
\end{figure}

\Fig{fig:input-granularity} compares incremental runtime across three input file granularities for Bagzel and Bagzel-xattr.

Across the evaluated range, input granularity has only a minor effect on incremental runtime. For both methods, runtime increases slightly as the number of input files grows (Bagzel: 42.67\,s up to 43.39\,s; Bagzel-xattr: 40.20\,s up to 40.78\,s from one to eight files), indicating limited sensitivity of incremental recomputation to file partitioning under fixed data volume.

Bagzel-xattr is consistently faster than Bagzel for all granularities, with gains of 2.47\,s, 2.53\,s, and 2.61\,s (5.8\%, 5.9\%, and 6.0\%, respectively), corresponding to a mean reduction of 5.9\% and a mean absolute gain of 2.53\,s on our system. The xattr benefit is stable across conditions and appears slightly larger at finer granularity.

Overall, the results indicate that incremental runtime increases only slightly as input granularity becomes finer. The benefit of server-side digest management also grows modestly with granularity. Within the evaluated range, both effects are small. However, for larger datasets with more input artifacts, the advantage of server-side digest management may become more pronounced, and input granularity may have a stronger impact on incremental rebuild performance.

\ifdefined\ArxivTexheightBalance
  \addtolength{\textheight}{\ArxivTexheightBalance}
\else
  \addtolength{\textheight}{-0.29cm}
\fi

\section{Conclusion}
This paper presented Bagzel, an artifact-based approach to robotics dataset construction that models preprocessing as a dependency graph with deterministic actions and cacheable artifacts. Across all evaluated execution modes, Bagzel outperformed a sequential rosbag2nuscenes baseline, with the largest gains in iterative workflows. On the 20.4\,GB dataset, Bagzel achieved up to \(386.26\times\) speedup in warm builds and \(7.21\times\) in incremental builds.

Scalability analysis over 5.1 to 20.4\,GB showed substantially lower runtime growth for Bagzel variants than for the baseline, especially in warm and incremental modes, where runtime remained nearly constant within the evaluated range. In addition, server-side digest management (Bagzel-xattr) provided consistent extra improvements over Bagzel, with a mean incremental runtime reduction of 5.9\% in the input granularity study. The benefit increased slightly with finer input granularity, suggesting that server-side digest management becomes more advantageous as the number of input artifacts grows.

These results indicate that artifact-based dataset construction is an effective strategy for reducing dataset update latency in robotics data pipelines. Although the evaluation focuses on runtime performance and is conducted on a single workstation with datasets ranging from 5.1 to 20.4\,GB, the findings provide initial evidence of the benefits of incremental artifact-based execution. Replication on larger datasets and distributed environments would further strengthen external validity.

Overall, modeling robotics dataset construction as an artifact-based build process substantially reduces recomputation overhead while enabling by design reproducible dataset generation through deterministic execution.





\section*{Acknowledgment}

The authors acknowledge support from BAAINBw and dtec.bw -- Digitalization and Technology Research Center of the Bundeswehr [project MORE]. dtec.bw is funded by the European Union -- NextGenerationEU. The authors thank the ROS and Bazel open-source communities.


\bibliographystyle{IEEEtran}
\bibliography{lepo,et_al}

\end{document}